\pdfoutput=1

\documentclass[11pt]{article}

\usepackage{ACL2023}

\usepackage{amsmath}

\usepackage{multirow}
\usepackage{graphicx}

\usepackage{times}
\usepackage{latexsym}

\usepackage[T1]{fontenc}

\usepackage[utf8]{inputenc}

\usepackage{microtype}

\usepackage{inconsolata}

\usepackage{url}

\title{PAI at SemEval-2023 Task 2: A Universal System for Named Entity Recognition with External Entity Information}

\author{
Long Ma\footnotemark[1], Kai Lu\footnotemark[1], Tianbo Che, Hailong Huang, Weiguo Gao, Xuan Li \\
\texttt{\{malong633,lukai113,chetianbo263,huanghailong590\}@pingan.com.cn} \\
\texttt{\{gaoweiguo801,lixuan208\}@pingan.com.cn}\\
  Ping An Life Insurance Company of China, Ltd.\\
}

\begin{document}
\maketitle
\renewcommand{\thefootnote}{\fnsymbol{footnote}}
\footnotetext[1]{These authors contributed equally to this work.} 

\begin{abstract}
%
%

The MultiCoNER II task aims to detect complex, ambiguous, and fine-grained named entities in low-context situations and noisy scenarios like the presence of spelling mistakes and typos for multiple languages. 
The task poses significant challenges due to the scarcity of contextual information, the high granularity of the entities(up to 33 classes), and the interference of noisy data. 
To address these issues, our team {\bf PAI} proposes a universal Named Entity Recognition (NER) system that integrates external entity information to improve performance. Specifically, our system retrieves entities with properties from the knowledge base (i.e. Wikipedia) for a given text, then concatenates entity information with the input sentence and feeds it into Transformer-based models. 
Finally, our system wins 2 first places, 4 second places, and 1 third place out of 13 tracks. The code is publicly available at \url{https://github.com/diqiuzhuanzhuan/semeval-2023}. 
\end{abstract}

\section{Introduction}

The objective of the MultiCoNER II shared task \cite{multiconer2-report} is to develop a robust named entity recognition (NER) system for multiple languages (including English, Spanish, Hindi, Bangla, Chinese, Swedish, Farsi, French, Italian, Portuguese, Ukranian, German and aforementioned mixed languages) that can perform well even when the input data has spelling errors, lacks contextual information, or contains out of knowledge base entities. This task entails the recognition of a total of 33 fine-grained entity types within six coarse-grained entity categories.
The entity types within each category are highly prone to confusion and ambiguity, and the testing data includes noise such as spelling errors that must be considered. 
Similar to the previous competition \cite{multiconer-report}, the data in this competition contains a lot of short and low-context sentences, which means the accurate prediction of entity types relies heavily on external knowledge. 

\begin{figure}[t]
\includegraphics[width=0.47\textwidth]{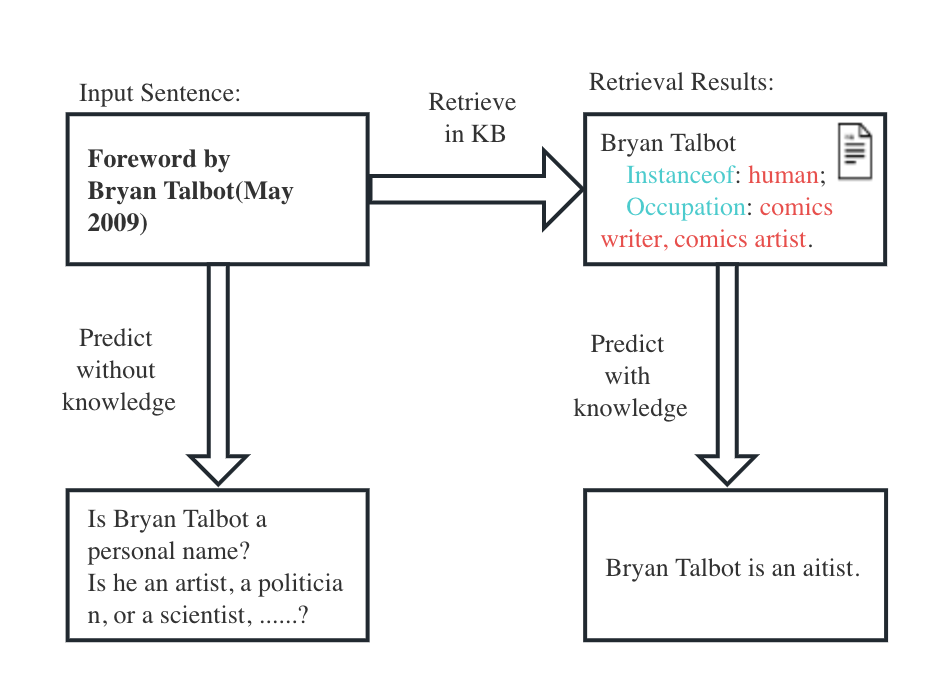}
\caption{A motivating example from the English dev set.
In the retrieval results, "Bryan Talbot" is an entity with two important properties. The blue phases are the property names and the red phases are the property values.}
\label{figure:motivation}
\end{figure}

Therefore, the key objective of our current work is to achieve effective integration of the model and external knowledge.
In this paper, extending our previous work \cite{ma-etal-2022-pai}, which has demonstrated the efficacy of the Dictionary-fused model, we propose a NER system based on Entity Property Knowledge Base. We constructed our knowledge base using all entities and their associated properties (including entity name, alias, sitelink title, description, instanceof, subclassof and occupation) from WikiData. 
In the retrieval module, we use string matching to retrieve entities with properties for a given sentence. Given the sequence length limitations of our model, we keep longer entities in sentences. Subsequently, we put the entity and property values into the external context in a specific format and feed the external context and original sentence into the NER model.
Figure \ref{figure:motivation} illustrates our motivation, demonstrating how the inclusion of property values can effectively improve performance in practice.

We further propose entity-aware attention mechanism, which can better model the semantic relationships between entities in sentences and external context than conventional attention. 
In the test phase, we use the voting strategy to form the final predictions.

We make the following observations according to our experiments:

\begin{enumerate}
    \item The system based on entity property knowledge base has significantly improved the effect. The knowledge base has a high entity coverage, and even more than 95 percent in some low-resource languages such as Bangla and Hindi.
    
    \item Different properties have different effects on different entity types. Specifically, occupation property can improve the recognition of fine-grained person entities.

    \item We have observed that the dictionary can greatly improve the performance on clean data, but may have a negative impact on noisy data. Knowledge retrieved by fuzzy matching such as ElasticSearch can help identify noisy entity.

    \item On some tracks, our entity aware attention can better capture the semantic relationships between entities in sentence and external context.
\end{enumerate}

\label{intro}

%
\section{Related Work}
Complex named entities (NE), like the titles of creative works, are not simple nouns and pose challenges for NER systems \cite{multiconer-data, multiconer2-data}. To mitigate these challenges, NER models utilizing external knowledge have achieved remarkable results.  Researchers have integrated gazetteers into models \cite{bender2003maximum, malmasi2016location} in earlier studies.
Recently, GEMNET \cite{meng2021gemnet} proposes a novel approach for gazetteer information integration, where a flexible Contextual Gazetteer Representation (CGR) encoder can be fused with any word-level model and a Mixture-of-Experts (MOE) gating network can overcome the feature overuse issue by learning to conditionally combine the context and gazetteer features, instead of assigning them fixed weights. GAIN \cite{chen-etal-2022-ustc} adapts the representations of gazetteer networks to integrate external entity type information into models. \cite{ma-etal-2022-pai} uses string matching to retrieve entities with types from the gazetteer and concatenates the entity information with the input text and feds it to models. However, the prerequisite for these approaches is to construct an entity dictionary with accurate types, which is extremely difficult to fulfill. Knowledge-fused methods mentioned in \cite{wang-etal-2022-damo, carik-etal-2022-su} build an information retrieval (IR) system based on Wikipedia to provide related context information to models. But these methods require retrieving high-quality context, which is difficult to guarantee in the MultiCoNER II shared task \cite{multiconer2-report}. 

Actually, Transformer-based models \cite{devlin-etal-2019-bert, NIPS2017_3f5ee243, yamada2020luke} have the ability to automatically learn the mapping between the entity properties and entity types on the basis of input sentence. Therefore, We propose a NER system that fuses entity dictionaries without accurate entity types to overcome these drawbacks of previous work.





\begin{figure*}[t]
\centering
\scalebox{0.9}{
\includegraphics[width=\textwidth]{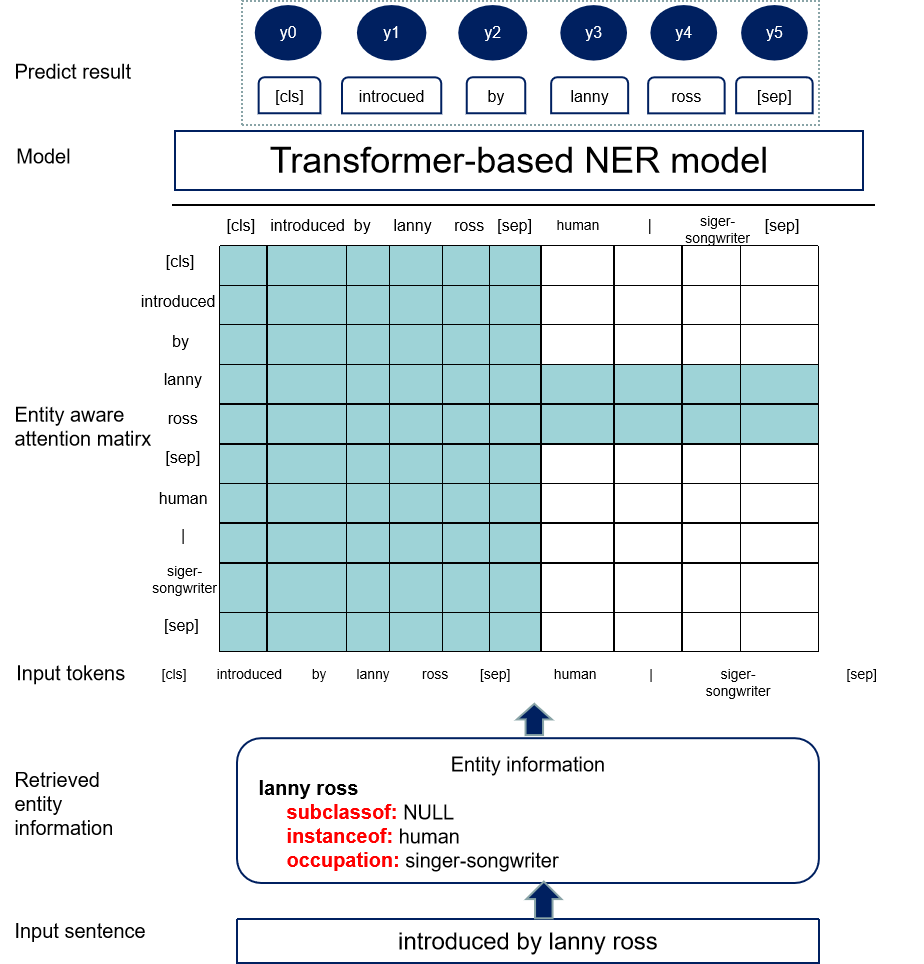}
}
\caption{The overall architecture of our proposed system shows a detailed case. "NULL" means that the field in the entity information is empty. In input tokens, different property values are separated by the token "|". The middle of the figure shows the specific entity aware attention matrix, which is described in section~\ref{subsection:attention}. The blue grid represents a value of 1, and the white grid represents a value of 0.}
\label{figure:architecture}
\end{figure*}

\begin{table*}[]
\centering
\begin{tabular}{|l|l|l|}
\hline
\textbf{Field}      & \textbf{Example "Victor Cousin"} & \textbf{Example "human"} \\ \hline
\textbf{Qid}        & Q434346                          & Q5                       \\ \hline
\textbf{Label}      & \textit{English: Victor Cousin}  & \textit{English: human}  \\ \hline
\textbf{Alias}          & \textit{English: NULL}          & \textit{English: [human being, humankind, ...]} \\ \hline
\textbf{Sitelink title} & \textit{English: Victor Cousin} & \textit{English: Human}                         \\ \hline
\textbf{Instanceof} & [Q5]                             & [Q55983715]              \\ \hline
\textbf{Subclassof} & NULL                             & [Q154954, Q164509, ...]  \\ \hline
\textbf{Occupation} & [Q4964182, Q82955, Q333634, ...] & NULL                     \\ \hline
\end{tabular}
\caption{Entity examples in WikiData. For the label, alias, and sitelink title fields, the examples only show the  value in English, but WikiData supports multiple languages. Qid is used to identify a unique entity. The value of the instanceof, subclassof, and occupation properties is qid. The value of label and sitelink title is a string, and other fields contain multiple strings. When the entity lacks a field, we use "NULL" value in the table.}
\label{table:wikientity}
\end{table*}

\begin{table*}[]
\centering
\begin{tabular}{|l|l|l|}
\hline
        & \textbf{Example "Victor Cousin"}            & \textbf{Example "human"}               \\ \hline
\textbf{Qid}     & Q434346                            & Q5                            \\ \hline
\textbf{Entity\_names}   & [Victor Cousin]                      & {[}human, human being, ...{]} \\ \hline
\textbf{Entity\_context} & human | philosopher | politician ...... & natural person | omnivore | mammal ......  \\ \hline
\end{tabular}
\caption{Entity names and context.}
\label{table:context}
\end{table*}

\section{Our System}
\label{section:system}

In this section, we introduce how our system works.
Given a sentence $x = \left \{ x_{0}, x_{1}, ..., x_{n-1}   \right \}$, $x$ consists of $n$ tokens, and this sentence is input to the knowledge retrieval module.
The retrieval module outputs the retrieved entities and property contexts. The system then concatenates the input sentence and outputs from the retrieval module and fed them to a Transformer-based model. To enhance the relevance of retrieved entities to the given information, we devised entity-aware attention mechanism. The overall architecture is shown in Figure~\ref{figure:architecture}.


\subsection{Knowledge Retrieval Module}

\paragraph{Knowledge Base Construction}
The entity property KB is constructed based on WikiData.
Wikidata contains more than 90 million different entities. 
Two entity examples with related fields are shown in Table~\ref{table:wikientity}.

$Qid$ represents an entity, and $Label_{qid}$, $Alias_{qid}$, $Sitelink\_title_{qid}$, $Subclassof_{qid}$, $Instanceof_{qid}$ and $Occupation_{qid}$ represents the values of different fields of the entity.
$Entity\_names_{qid}$ represents the entity names, consists of $Label_{qid}$, $Sitelink\_title_{qid}$ and $Alias_{qid}$.
$Property\_qid_{qid}$ represents the property values in qid format, consists of $Subclassof_{qid}$, $Instanceof_{qid}$ and $Occupation_{qid}$.
$Property\_names_{qid}$ represents the property names in text format, which is the label field values of $Property\_qid_{qid}$.
$Context_{qid}$ represent the entity context. 
In $Entity\_context_{qid}$, $Property\_names_{qid}$ is separated by a special token "\textit{\textbf{|}}".
Two entity examples with names and context are shown in Table~\ref{table:context}.
The knowledge base contains all entity names and contexts.
%

\begin{table}[t]
\begin{tabular}{ccc}
\hline
\textbf{Language} & \textbf{Manual Dict} & \textbf{Property Dict}  \\ \hline
BN & 41.7 & 96.9 \\ \hline
DE & 55.6 & 98.9 \\ \hline
EN & 60.6 & 94.3 \\ \hline
ES & 57.0 & 95.1 \\ \hline
FA & 34.6 & 84.8 \\ \hline
FR & 53.5 & 95.2 \\ \hline
HI & 37.0 & 99.6 \\ \hline
IT & 60.4 & 96.2 \\ \hline
PT & 56.3 & 93.7 \\ \hline
SV & 55.9 & 93.9 \\ \hline
UK & 26.9 & 65.2 \\ \hline
ZH & 42.9 & 99.6 \\ \hline
AVG & 48.5 & 92.8 \\ \hline
\end{tabular}
\caption{ Coverage rate on dev data. The definition of coverage rate is the same as \cite{chen-etal-2022-ustc}. "Coverage rate" refers to the number of entities in the data also found in our dictionary / the number of entities in the data.}
\label{table:cover}
\end{table}

\begin{table*}[]
\centering
\scalebox{0.9}{
\begin{tabular}{cccccccccccccc}
\hline
 \textbf{System} & \textbf{DE} & \textbf{BN} & \textbf{ZH} & \textbf{EN} & \textbf{ES} & \textbf{HI} & \textbf{SV} & \textbf{FA} & \textbf{FR} & \textbf{IT} & \textbf{PT} & \textbf{UK} & \textbf{AVG}         \\ \hline
\textbf{Baseline}                      & 72.1 & 75.0 & 68.3 & 69.7 & 73.4 & 71.1 & 75.2 & 60.2 & 74.0 & 78.5 & 71.1 & 67.6 & 71.3   \\ \hline
\textbf{Manual Dict}        &    78.1 & 78.5 & 75.2 & 74.5 & 73.8 & 79.2 & 78.8 & 66.7 & 77.9 & 78.7 & 77.5 & 69.2 & 75.7           \\ \hline
\textbf{Property Dict} &   \textbf{88.5} & \textbf{89.4} & \textbf{91.8} & \textbf{81.2} & \textbf{81.9} & \textbf{88.1} & \textbf{84.2} & \textbf{73.5} & \textbf{83.2} & \textbf{85.3} & \textbf{80.6} & \textbf{74.8} & \textbf{83.5}   \\ \hline
\end{tabular}
}
\caption{ Dev micro F1 score with different dictionary. \textbf{Manual Dict} means the model with type dict which needs manual relational mapping, \textbf{Property Dict} means the model with entity property knowledge. }
\label{table:main}
\end{table*}

\begin{table*}[t]
\centering
\scalebox{0.9}{
\begin{tabular}{cccccccccccccc}
\hline
 \textbf{Data Type} & \textbf{DE} & \textbf{BN} & \textbf{ZH} & \textbf{EN} & \textbf{ES} & \textbf{HI} & \textbf{SV} & \textbf{FA} & \textbf{FR} & \textbf{IT} & \textbf{PT} & \textbf{UK}        \\ \hline
\textbf{Clean Data}           & 88.09 & 84.39 & 86.23 & 86.16 & 79.35 & 80.96 & 81.53 & 68.46 & 89.5 & 88.94 & 84.56 & 71.28  \\ \hline
\textbf{Noisy Data}  &   -- & -- & 41.9 & 65.41 & 55.25 & -- & 55.22 & -- & 78.71 & 76.53 & 76.12 & --      \\ \hline
\textbf{Overall} &  88.09 & 84.39 & 74.87 & 80 & 71.67 & 80.96 & 72.38 & 68.46 & 86.17 & 84.88 & 81.64 & 71.28   \\ \hline
\end{tabular}
}
\caption{  Macro F1 score on clean data and noisy data in the evaluation phase. "--" means that there is no noisy data. }
\label{table:noisy}
\end{table*}

\begin{table}[t]
\centering
\begin{tabular}{cc}
\hline
 & \textbf{Average F1}  \\ 
 \hline
All property & \textbf{83.5} \\ 
\hline
All property - subclassof & 82.5 \\ 
\hline
All property - instanceof &  80.4 \\ 
\hline
All property - occupation & 79 \\ 
\hline
\end{tabular}
\caption{ Average micro F1 score of all monolingual tracks in the dev set when using different properties to construct the dictionary. }
\label{table:property-main}
\end{table}

\begin{table}[h]
\begin{tabular}{ccc}
\hline
 & \textbf{BN} & \textbf{IT} \\ 
 \hline
 \textbf{Model with default attention}   & 78.5 & 78.7 \\ 
 \hline
\textbf{Model with  entity}  & 79.5 & 79 \\ 
\hline
\end{tabular}
\caption{ Dev micro F1 score with entity aware attention on Bangla and Italian. }
\label{table:span}
\end{table}

\paragraph{Entity Retrieval}
Given an input sentence, the retrieval module retrieves entity names by string matching. The external context consists of matched entity names and corresponding entity contexts. We prioritize preserving the longer entities when encountering overlapping matches or the need to truncate information.
Finally, retrieval module returns multiple entity context pairs, $pairs = \left [ (entity_0, context_0), (entity_1, context_1), ...\right ]$. The length of the $pairs$ is $m$. 

\subsection{Entity Aware Attention}
\label{subsection:attention}

Our system concatenates input sentence $x$ and entity contexts together, and entity contexts are separated by a special token "\textbf{\$}".

%

In order to better use the external knowledge in Transformer-based model, entity-aware attention is proposed to link entities in sentence with corresponding contexts.
Entity-aware attention matrix $M$ is a binary matrix with values 0 or 1, which makes context only attended by the corresponding entity. Under both conditions $M_{i,j} = 1$, other times $M_{i,j} = 0$.
In the first condition, $i, j < n + 2$. $n+2$ takes into account [CLS] token at the front of the sentence and [SEP] token at the end of the sentence.
In the second condition, i in $position_{entity_k}$, and j in $position_{context_k}$, $0 <= k < m$.
$position_{entity_k}$ and $position_{context_k}$ are position subscript sets of $entity_k$ and $context_k$ respectively.
Entity-aware attention matrix $M$ is input into Transformer-based NER model as attention mask.
An entity-aware attention matrix example is shown in Figure~\ref{figure:architecture}.


\section{Exprimental Setup}
\label{section:setting}
\subsection{Data and Evaluation Metrics}
%
MultiCoNER II contains 6 coarse-grained labels, which can be further subdivided into 33 fine-grained labels. The specific labels are shown in table~\ref{table:property}.
Fine-grained named entity recognition (NER) is a particularly challenging task, especially in low-context and ambiguous situations.
For instance, identifying names of individuals who are scientists or musicians can be super difficult without sufficient context.

The task has 12 monolingual tracks, including multiple low-resource language tracks that are difficult to obtain external knowledge, such as Hindi and Bangla.
In this paper, all results use entity level micro F1 as evaluation metric except table~\ref{table:noisy}.
Table~\ref{table:noisy} use macro F1 which is official evaluation metric.

%

\subsection{Training detail}

We use Chinese BERT with Whole Word Masking \cite{chinese-bert-wwm} as our pre-trained model in Chinese and utilize BERT multilingual base model (uncased)\cite{wolf-etal-2020-transformers} in other tracks.
An 8-fold cross-validation training strategy is applied in the evaluation except for MULTI task.
In the evaluation phase, all the best models vote on the prediction results, and the voting weights are determined by the F1 score of each model on the validation set.

\section{Results and Analysis}
\label{section:results}

\subsection{Main Results}
To show the effectiveness of our system, we evaluate the results of a baseline system without knowledge.
We reproduce the dictionary construction process of \cite{chen-etal-2022-ustc} on MultiCoNER II, and obtain a type dictionary through manual relational mapping.
The results of the baseline, Manual Type dictionaries, and our system are shown in Table~\ref{table:main}.
The automatic property dictionary exceeds the baseline by 12.2 F1 on average, demonstrating the effectiveness of our system.
Without manual relational mapping, our proposed property dictionary outperforms the manually constructed type dictionary by 7.8 F1 on average, demonstrating the capability of the model to learn the correlation between property values and entity types automatically.

\subsection{Coverage Rate Trial}

Previous studies\cite{rijhwani-etal-2020-soft, meng2021gemnet} have found that systems with higher entity coverage have higher performance.
Table~\ref{table:cover} shows the entity coverage of Manual Type Dictionary and Property Dictionary.
The coverage rate of the property dictionary is 44 percent higher than that of the artificial dictionary and even reaches 99 percent in some languages, such as Chinese and Hindi.
The track with higher coverage rate has bigger improvement in Table~\ref{table:main}.

\begin{table*}[t]
\centering
\scalebox{0.9}{
\begin{tabular}{c|ccccc}
\hline
 
\begin{tabular}{c}
     \textbf{Coarse-grained}  \\
     \textbf{taxonomy}
\end{tabular}  
  &
\begin{tabular}{c}
     \textbf{Fine-grained}  \\
     \textbf{taxonomy}
\end{tabular}  
  & \textbf{All property} & \begin{tabular}{c}
     \textbf{All property}  \\
     \textbf{- subclassof}
\end{tabular}  
& \begin{tabular}{c}
     \textbf{All property}  \\
     \textbf{- instanceof}
\end{tabular}   
& \begin{tabular}{c}
     \textbf{All property}  \\
     \textbf{- occupation}
\end{tabular}    
\\ \hline
\multirow{5}{*}{Creative Works}  & ArtWork & 76.8 & 76.2 & \textbf{69.8} & 78.7 \\ \cline{2-6}
 ~ & MusicalWork & 82.4 & 81.4 & \textbf{77.3} & 82.6 \\ \cline{2-6}
 ~ & Software & 82.0 & 81.5 & \textbf{79.2} & 84.0 \\ \cline{2-6}
 ~ & VisualWork & 85.3 & 83.9 & \textbf{81.2} & 87.6 \\ \cline{2-6}
 ~ & WrittenWork & 79.7 & 78.3 & \textbf{73.8} & 77.6 \\ \hline
\multirow{7}{*}{Group}  &  PublicCorp & 80.7 & 81.6 & \textbf{66.5} & 78.3 \\ \cline{2-6}
 & AerospaceManufacturer & 92.0 & 91.1 & \textbf{83.5} & 87.4 \\ \cline{2-6}
 & CarManufacturer & 84.0 & 82.8 & \textbf{80.7} & 82.4 \\ \cline{2-6}
 & MusicalGRP & 90.5 & 89.3 & \textbf{85.8} & 90.5 \\ \cline{2-6}
 & ORG & 79.1 & 78.6 & \textbf{68.6} & 76.9 \\ \cline{2-6}
 & PrivateCorp & 83.3 & 74.8 & \textbf{70.3} & 75.5 \\ \cline{2-6}
 & SportsGRP & 89.8 & 87.5 & \textbf{86.9} & 89.8 \\ \hline
\multirow{4}{*}{Location} & Facility & 76.6 & 75.5 & \textbf{73.2} & 79.0 \\ \cline{2-6}
 & HumanSettlement & 90.7 & 90.2 & \textbf{87.7} & 90.3 \\ \cline{2-6}
 & OtherLOC & 74.3 & \textbf{66.5} & 71.1 & 73.3 \\ \cline{2-6}
 & Station & 82.5 & 82.9 & \textbf{77.4} & 84.0 \\ \hline
\multirow{5}{*}{Medical} & AnatomicalStructure & 75.7 & 78.2 & \textbf{72.0} & 74.8 \\ \cline{2-6}
 & Disease & 77.2 & \textbf{72.5} & \textbf{72.5} & 74.8 \\ \cline{2-6}
 & MedicalProcedure & 73.4 & \textbf{68.0} & 68.6 & 76.8 \\ \cline{2-6}
 & Medication/Vaccine & 81.9 & 79.4 & \textbf{78.4} & 80.8 \\ \cline{2-6}
 & Symptom & 82.1 & 74.9 & \textbf{70.0} & 76.6 \\ \hline
\multirow{6}{*}{Person} & Artist & 90.3 & 89.9 & 89.9 & \textbf{82.7} \\ \cline{2-6}
 & Athlete & 87.1 & 88.6 & 87.6 & \textbf{74.1} \\ \cline{2-6}
 & Cleric & 81.5 & 78.9 & 81.6 & \textbf{65.2} \\ \cline{2-6}
 & OtherPER & 75.6 & 75.0 & 73.4 & \textbf{57.5} \\ \cline{2-6}
 & Politician & 83.6 & 83.7 & 84.1 & \textbf{62.8} \\ \cline{2-6}
 & Scientist & 77.7 & 78.9 & 77.0 & \textbf{52.7} \\ \cline{2-6}
 & SportsManager & 89.0 & 90.5 & 89.9 & \textbf{67.9} \\ \hline
\multirow{5}{*}{Product} & Clothing & 65.1 & \textbf{59.2} & 65.5 & 62.3 \\ \cline{2-6}
 & Drink & 77.8 & \textbf{71.1} & 75.3 & 78.6 \\ \cline{2-6}
 & Food & 66.0 & \textbf{61.4} & 63.4 & 66.0 \\ \cline{2-6}
 & OtherPROD & 73.0 & \textbf{68.0} & 70.2 & 72.8 \\ \cline{2-6}
 & Vehicle & 69.0 & \textbf{65.5} & 70.4 & 67.6 \\ \hline
\end{tabular}
}
\caption{ The F1 score on taxonomy with different properties combination. In the table, the bold numbers are the lowest F1 on taxonomies. }
\label{table:property}
\end{table*}

\subsection{Effect of Property}
In order to analyze the impact of different properties on the results, we conduct ablation experiments of properties. 
Table~\ref{table:property-main} shows the results of using different properties.
We observe that the \textit{Occupation} property has a greater impact on the performance. Without \textit{Occupation} property, the average results drop by  4.5 F1.
Further, we analyze the relationship between properties and taxonomies. Table~\ref{table:property} shows the taxonomy F1 score when using different property combinations.
The \textit{subclassof} property is strongly related to product class. 
The \textit{instanceof} property is strongly correlated with Creative Works, Group, Location, and Medical classes.
\textit{Occupation} is a property about people. Experiments show that occupation property plays a crucial role in identifying Person class.

\subsection{Effect of Entity Aware Attention}

We propose entity-aware attention, which can represent the relationship between entities and context in a sentence.
Table~\ref{table:span} shows the effect of entity-aware attention on several languages.
The entity-aware attention improves by 1 F1 and 0.3 F1 over the baseline on Bangla and Italian, demonstrating its effectiveness.

\subsection{Clean Data and Noisy Data}

Table~\ref{table:noisy} shows the results of our system on clean data, noisy data, and overall data in evaluation.
The results demonstrate that our system achieves higher performance on clean data than on noisy data, which poses a challenge due to the noisy entities that can not be retrieved through string matching. The Chinese track exhibits the largest discrepancy between clean and noisy data since the Chinese training and validation sets only contain clean data, while the test set includes both clean and noisy data.







\section{Conclusion}
\label{section:conclusion}

In this paper, we describe our NER system based on entity properties, which wins two tracks in MultiCoNER II shared task.
We construct a KB based on entity properties, which is used to retrieve the relevant entity names and contexts for a given sentence.
Our property dictionary is built without the need for manual relational mapping and achieves high coverage on the test set.
We have found that different entity types require different properties.
We propose the entity-aware attention mechanism to better learn the relationship between entities and contexts.
In the future, we plan to adopt fuzzy matching to improve the performance on noisy data and explore our system for other low-resource tasks.




\bibliography{anthology,custom}
\bibliographystyle{acl_natbib}



\end{document}